# Spatial Feature Extraction in Airborne Hyperspectral Images Using Local Spectral Similarity

Anand S Sahadevan, Arundhati Misra, Praveen Gupta
Space Applications Centre, ISRO, Ahmedabad-380015

*Abstract*- Local spectral similarity (LSS) algorithm has been developed for detecting homogeneous areas and edges in hyperspectral images (HSIs). The proposed algorithm transforms the 3-D data cube (within a spatial window) into a spectral similarity matrix by calculating the vector-similarity between the center pixel-spectrum and the neighborhood spectra. The final edge intensity is derived upon order statistics of the similarity matrix or spatial convolution of the similarity matrix with the spatial kernels. The LSS algorithm facilitates simultaneous use of spectral-spatial information for the edge detection by considering the spatial pattern of similar spectra within a spatial window. The proposed edge-detection method is tested on benchmark HSIs as well as the image obtained from Airborne-Visible-and-Infra-Red-Imaging-Spectrometer-Next-Generation (AVIRIS-NG). Robustness of the LSS method against multivariate Gaussian noise and low spatial resolution scenarios were also verified with the benchmark HSIs. Figure-of-merit, false-alarm-count and miss-count were applied to evaluate the performance of edge detection methods. Results showed that Fractional distance measure and Euclidean distance measure were able to detect the edges in HSIs more precisely as compared to other spectral similarity measures. The proposed method can be applied to radiance and reflectance data (whole spectrum) and it has shown good performance on principal component images as well. In addition, the proposed algorithm outperforms the traditional multichannel edge detectors in terms of both fastness, accuracy and the robustness. The experimental results also confirm that LSS can be applied as a pre-processing approach to reduce the errors in clustering as well as classification outputs.

*Keywords*- Hyperspectral image, Spectral distance measures, Local spectral similarity, Classification, Clustering

## 1. Introduction

Hyperspectral sensors simultaneously collect spatial-spectral information over a large area of a ground-based scene and produce hundreds of contiguous bands. Each pixel in the resulting hyperspectral image (HSI) contains a sampled spectral measurement of radiance, which can be further processed to form the corresponding surface reflectance to identify the material present in the scene. Reflectance response of the surface mainly depends on the properties and the composition of the material in the scene. Certain materials have high reflectance responses over a broad wavelength range and form bright and sharp features in the corresponding bands of the image cube. Conversely, certain substances have a peak reflectance response over a very narrow wavelength range and form sharp features, which are seen only in a few bands. In general, the variability in spectral response of sharp features over different spectral bands introduce inconsistent edges and homogeneous regions in multiple bands.

Detection of similar land patches and their boundaries are one of the major tasks in both airborne and space borne image analysis (e.g. classification, segmentation and clustering). Accuracy of these techniques can be enhanced by identifying the homogeneous regions (or their edges) within the scene [25] as a preprocessing step. In specific, the inconsistent class statistics obtained due to the outliers (pixel near to the edges) can be prevented by ensuring that the pixels are extracted from homogeneous regions. In fact, the regions with high spatial-spectral similarity are associated with finding those areas that give low response to local edge detectors. Accordingly, while extracting edges, low values in edge map indicate homogeneity and high value indicate spatial-spectral dissimilarity or edge features.

Multichannel edge detection techniques can be grouped into three categories such as monochromatic approaches, vector based approaches and feature-space based approaches. Monochromatic methods process each multichannel component separately and combining (image fusion or multidimensional gradient) the individually gained results into a single edge map [12],[8]. However, in HSI, edge features may be observable over a narrow band, producing different spectral bands with inconsistent edges. Feature-based edge-detection methods have also been used for detecting edges in colour images and multispectral images (MSI) [15],[13]. In [15], subspace classification on multivariate features have been utilized to determine edge pixels in colour images. Dinh et. al. [9] has developed a gradient-feature-clustering technique to detect edges in MSIs. Moreover, researchers have also examined the capability of spatial-spectral feature space for detecting edges in HSIs. However, the lack of suffiecent number of ground truth samples and localization error in the groundtruth boundaries may lead to erroneous edge/homogeneous feature detection in feature-space based methods. Vector based approaches [4] preserve the vector nature of HSI throughout the computation and use various features (curvature properties of vector fields, vector order, vector similarity etc.) of n-dimensional vector space to detect edges in HSI. Machuca and Phillips [18] proposed a vector-based method that uses rotational and curvature properties of vector fields to identify edges. Vector difference based approaches [12] were also studied for replacing gray level differences of adjacent pixels by vector differences. Trahanias and Venetsanopoulos [28] developed vector-order-statistics for detecting edges from colour images.

HSIs usually consist of hundreds of contiguous spectral bands. However, spectral mixtures or edges that appear in a small subset of bands and inconsistent edges from narrow channels may produce low spatial-spectral homogeneity. This issue makes the edge detection more challenging in HSIs. To tackle these challenges, it is critical to consider the spatial relationships between spectral vectors as well as spectral information distributed over the entire bands. Therefore, In this paper, local spectral similarity (spectral similarity with in a neighborhood) based edge-detection method for HSIs is proposed. The proposed algorithm transforms the 3-D data cube (within a neighborhood) into a 2-D spectral similarity matrix by calculating the similarity between the center pixel-spectrum and the neighborhood spectra. As a result, abrupt changes in the similarity matrix represents the changes in the local spatial extent. The final edge intensity is derived upon statistical accumulation of spectral similarity information contained in the similarity matrix.

The key distinctive mark of the proposed algorithm is that it is simple to implement and the simultaneous use of spatial and spectral information may reveal pixels with anomaly/unique spectrum (compared to other spectra with in a neighborhood), enhance different level of homogeneity, etc. This is best demonstrated when mean of the similarity matrix is applied to find the intensity of edges (mean is prone to outliers). Since LSS can be applied to find the homogeneous areas, it can be applied as a pre-processing approach before endmember extraction to mask the pixels with mixed responses or anomalies (the edges encompassing the mixed pixels in the scene have responses greater than zero). i.e., masking out the edges prior to the endmember detection, clustering and classification algorithm can improve their results.

This paper is organized as follows. In Section 2 we present the formulation of the proposed LSS based edge detection algorithm and its implementation for HSI. In Section 3, we discuss different spectral similarity measure, different criteria for evaluating the performance of edge detection and the application of edge detection for improving clustering and classification accuracies. In Section 4 we present results of applying the algorithm to real data from the benchmark AVIRIS dataset as well as AVIRIS-NG image. In this section we compare the performance of different spectral similarity method and their robustness to multivariate Gaussian noise, spatial resolution and dimensionality reduction as well. Our conclusions are presented in Section 5.

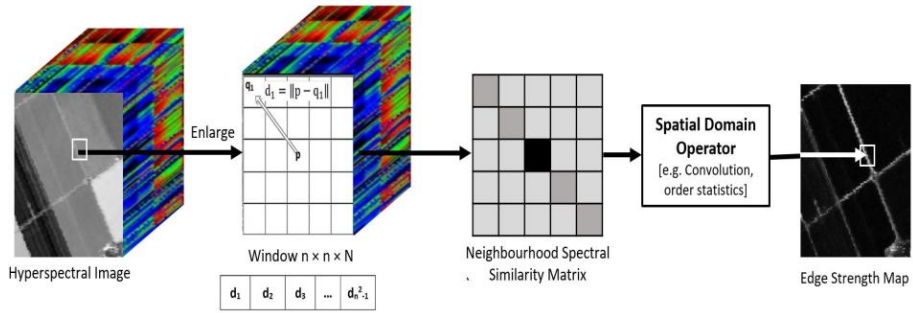

**Figure 1**. Block diagram of the proposed Local spectral similarity (LSS) approach. Black pixel represents the center pixel (excluded from point statistics computation)

## 2. Problem Formulation

Edges (or spatial disparities) are image features that are widely used in image analysis to outline the boundaries of objects. An edge is defined as a local change or discontinuity in image luminance [1]. Therefore, identification of spatial discontinuities is critical to identify spatially homogeneous regions (or edges) in images. Generally, in edge enhancement, spatial discontinuities (or disparities) in a grayscale image are enhanced by some spatial operator. However, in multispectral images, every single band (uncorrelated bands) provide different information to enable an accurate representation of image objects. Meanwhile, in HSIs, weak edge features with in a subset of bands and strong correlation between adjacent bands may reduce the spatial disparity among adjacent bands. In such cases, edge enhancement may introduce inconsistent edges in multiple bands.

In HSIs (Atmospherically corrected surface-reflectance data), a single pixel-vector (a stack of pixels belong to different bands) at a location (i, j) in the image cube represents a reflectance spectrum corresponds to a target material. In HSIs, particularly in the homogeneous region, spatial correlation exists due to the similarity between the spectra in the close proximity. However, the similarity among those pixel-vectors may decrease mainly due to the spatial disparities (edges, noise, etc.). Therefore, one of the important tasks in HSI analysis is to find the locations at which the image undergoes considerable spatial variations. Furthermore, HSIs suffer from three drawbacks in edge (or homogeneous area) detection. These include noise in the image, weak spectral features (or weak edges) that appear in narrow bands, and the occurrence of inconsistent edges at different narrow bands. To tackle these issues, it is necessary to consider the spatial-spectral relationships between neighborhood vectors. Therefore, computing a local variation of pixel-vectors within a neighborhood window may be more appropriate for HSIs [9]. From the applications point of view, the task of edge detection is linked with finding homogeneous areas in the HSIs. i.e., edges or spatial disparity may create ambiguities, which may result in erroneous outcomes in clustering, segmentation, end-member extraction and classification of HSIs. Therefore, excluding the edges (or dissimilar pixel) before applying any algorithm may improve the performance of the above mentioned methods.

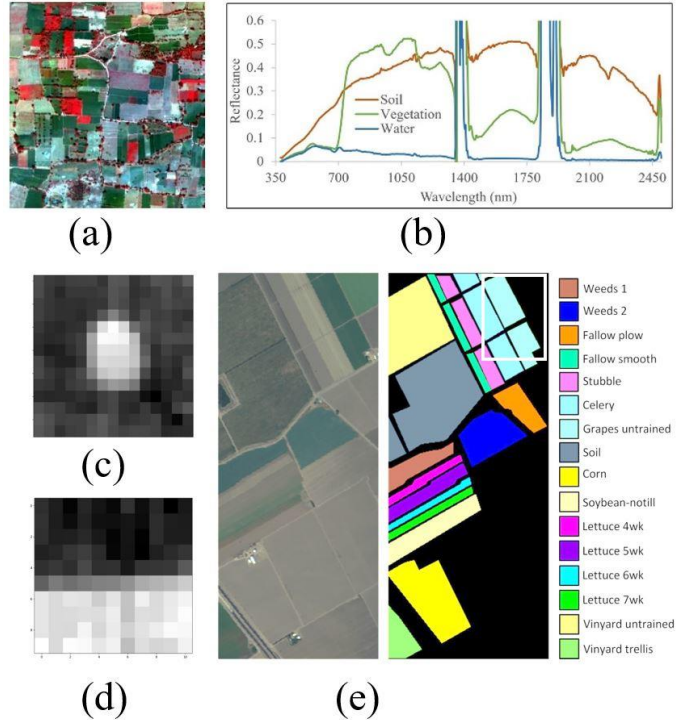

**Figure 2**: (a) AVIRIS-NG image (false colour composite) captured over agricultural field at Muddur, Karnataka, India. (b) Reflectance spectra obtained from the image. (c) spatial subset (AVIRIS-NG image, 805 nm) shows a tree (bright patch in the middle) and the agricultural land. (d) spatial subset (AVIRIS-NG image, 805 nm) shows the cropping land (bright patch), the bare soil (dark patch) and the boundary between the patches.(e) shows the true color composite of the Salinas image, the corresponding ground truth data and the white box shows the subset of the image used for testing the performance of edge detection methods

It is known that discontinuities in textured image represents a critical feature called boundaries. Therefore, boundary detection plays and important role in texture classification and is an active topic of research in pattern recognition. The main challenge of texture classification is to deal with the rotation, illumination, and scale. Accordingly, descriptors (e.g. local binary pattern (LBP); Scale Invariant Feature Transform (SIFT) etc.) have been proposed to describe patches within the image to deal with the variations in grayscale images [30]. LBP has been applied to many application areas, such as texture recognition, edge detection etc. In LBP, a local region can be originally characterized by a $p$-dimensional difference vector $\vec{d}_p$ between the central pixel $g_c$ and its neighbors $g_p$, where $\vec{d}_p = g_1 - g_c, g_2 - g_c, \ldots, g_p - g_c$ and $p$ is number of neighbors. The LBP descriptor has limited capability to capture more discriminative information because only the sign of the difference vector $d$ is used to represent the local region. The magnitude of difference vector $d$, which also contains discriminative information is completely discarded. As a result, local regions with different levels of grayscale difference may have the same LBP codes and lead to the misclassification of these different patterns. To overcome this shortcoming, Guo et. al. [11] proposed a method, which combines three components of the difference vector $d$ as: the sign descriptor, the magnitude descriptor and the central pixel descriptor. However, in HSI, single pixel that represents a target spectrum (distinct spectral responses over different wavelengths) limits the direct use of sign descriptor and the center pixel descriptor. Therefore, we propose a similarity based local feature descriptor that utilizes the vector similarity measures to reflects the strength of relationship (or similarity) between pixel-vectors in a neighborhood.

## 2.1 Local Spectral Similarity(LSS) based edge detection

In [11], a local region is denoted by the difference vector $d$ and this vector is decomposed into three components: the sign descriptor, the magnitude descriptor and the central pixel descriptor. However, in HSI, pixels intensities across different bands form a target spectrum. Therefore, computing the difference vector is not practical in spectral domain. Based on this consideration, we propose to use a similarity based local feature (Local Spectral Similarity, LSS) descriptor for HSI for defining the homogeneous patches [21]. In the LSS descriptor (see Fig. 1), magnitude of the similarity is described by spectral similarity measures (e.g. Euclidean distance), then the edge magnitude information obtained from the spectral-similarity matrix $d_p$ is extracted using the spatial operator (e.g. spatial convolution or order statistics) defined over some neighborhood of the center pixel.

$$y[i,j] = \sum_{u=-k}^{k} \sum_{v=-k}^{k} d[u,v] \cdot h[i-u, j-v] \tag{1}$$

where, $d[u,v]$ is the spectral similarity matrix within a spatial window and $h$ represents the convolution kernel (e.g. mean filter kernel). We have also used ordered statistics such as median, minimum, maximum and mid point, median absolute deviation (MAD) to compute point measure of the similarity matrix. These point statistical measures identify the distribution of spectrally similar pixels within a spatial window. However, the degree of similarity may vary depending on the spectral-similarity measure. In a homogeneous region, these measure will produce very low value because of the high similarity between the center pixel spectrum and the neighborhood spectra. On the contrary, in a high disparity region, the spectral distance measure will produce high values due to the high dissimilarity between the center pixel spectrum and the neighborhood spectra. More specifically, the point statistical measures can be used as a point feature for representing the uniformity (or non-uniformity) with in a spatial window. They can mitigate the influence of rotation, illumination or noise, different edge-strength across multi-channels and demonstrate better robustness for detecting boundaries in the HSI. Moreover, extremely noisy bands can also be removed prior to the spectral similarity measurement to avoid influence of high noise levels.

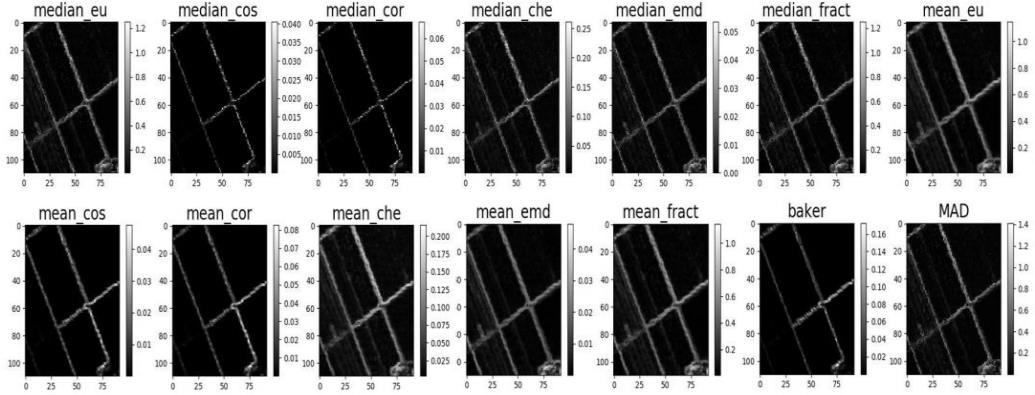

**Figure 3.** Edge maps generated using LSS (3×3 window) approach (combinations of different spectral similarity measures and the point statistical measures)

## 2.2 LSS in higher-dimensional space

The spectral information in HSI is usually seen in small spectral absorption features that are superimposed in the highly collinear bands. In HSI, with hundreds of wavebands, the required number of pixels to address the spectral variability grows exponentially (Hughes phenomenon). Therefore, the curse of dimensionality becomes one of the major obstacles for data mining techniques such as similarity indexing, search and retrieval, classification etc. Therefore, selecting the appropriate distance measures for finding the spectral similarity is critical to derive edges from the HSIs.

In HSIs, spatial-spectral similarity can be linked to the vector similarity (spectral similarity) with in a neighbourhood. Moreover, vector based edge detection methods can preserve the vector nature of HSIs and perform edge detection in vector space. Detecting edges in vector-space [28] has shown great potential and is used in state-of-the-art color image processing techniques. The typical image neighborhood consists of the closest four or eight neighbors to the centre pixel. Therefore, finding the similarity between these neighborhood vectors using vector-similarity measures is a key task to perform edge detection. However, for such applications, curse of high dimensionality and band-collinearity tend to be a major obstacle in the development of detecting spectral disparity. In addition to this, Bayer et al. [5] has shown that under certain reasonable assumptions the ratio of the distance between the nearest and the farthest neighbors to a given target in higher dimensional space is almost one for a variety of distance functions. In such cases, the contrast between data points may be poor. Therefore, the choice of spectral distance measure is more important for detecting the spectral similarity.

Keeping such a view, we examine usefulness of the widely used distance measures for computing the spatial-spectral similarity in a local neighborhood. In this study, we explored the usefulness of widely used spectral similarity measures such as Euclidean (EU), Manhattan distance (MAN), Fractional distance measure (FRACT)[2], Chebyshev distance (CHE)[7], Cosine similarity (COS) [22], Spectral information divergence (SID)[22], Correlation distance (COR) [7], Earth movers distance (EMD) [3] etc. The $L_k$ norm distance function can be defined as,

$$d_k(g_p, g_c) = \sum_{i=1}^{N} ((\|g_p^i - g_c^i\|^k)^{1/k}) \qquad (2)$$

$L_1$ metric is called as Manhattan distance and $L_2$ is called as Euclidean distance metric. Aggarwal et. al. [2] illustrated the meaningfulness of $L_k$ norm worsen faster with increase in dimensionality for higher values of $k$. Authors also suggested that lower values of $k$ ($k = 1$ or 2) is preferable to use for a problem with fixed dimensionality (high). Similarly, we computed, fractional distance measure ($0 < k < 1$), in which $k$ is allowed to be a fraction less than 1. Correlation distance is defined as,

$$1 - \frac{(\vec{g}_p - \bar{g}_p) \cdot (\vec{g}_c - \bar{g}_c)}{\|\vec{g}_p - \bar{g}_p\|_2 \|\vec{g}_c - \bar{g}_c\|_2} \quad (3)$$

where, $\bar{g}_p$ and $\bar{g}_c$ show the mean of the elements of $\vec{g}_p$ and $\vec{g}_c$ respectively. In Eq.3, the correlation measure is converted into the distance measure for finding the spectral-similarity.

## 2.3 LSS in Projected Space (lower-dimensional space)

In HSI, because of the high spectral sampling (or spectral-resolution), information in adjacent bands is highly correlated. This inter-band correlations may lead to lower dimensional space spanned by the data [17]. Therefore, several dimensionality reduction (DR) methods have been developed for the exploitation of the linear (e.g. Principal component analysis, PCA) as well as non-linear (e.g. locally linear embedding) characteristics of the HSI [27]. Given a data matrix $X$, the dimensionality reduction problem seeks to find a set of coordinates $Y = \{y_i\}_{i=1}^n, y_i \in \mathbb{R}^p$, where typically, $m << p$ through a feature mapping $\Phi: x \rightarrow y$, which may be linear or nonlinear. Manifold learning algorithms such as isometric feature mapping (ISOMAP)[17], locally linear embedding (LLE) [24], Spectral Embedding (SE) and Multi dimensional Scaling (MDS) have received much attention in HSI data analysis domain.

Linear dimensionality reduction methods perform rotation and scaling on the data and the data projection is performed using a transformation matrix computed using data variance or discriminant features etc. However, in non-linear dimensionality reduction, the global manifold learning methods assume that the local feature space formed by the nearest neighbors is linear, and the global nonlinear transformation can be found by connecting these piecewise linear spaces. Local kernel-based manifold-learning methods are initiated by constructing a nearest neighborhood for each data point, and the local structures are then used to obtain a global manifold. Keeping such a view, we have also explored the utility of global as well as local neighborhood preserving transformation for directly extracting homogeneous areas and boundaries in the image. Moreover, LSS is applied on lower-dimensional space to explore the potential of dimensionality reduction methods for detecting edges.

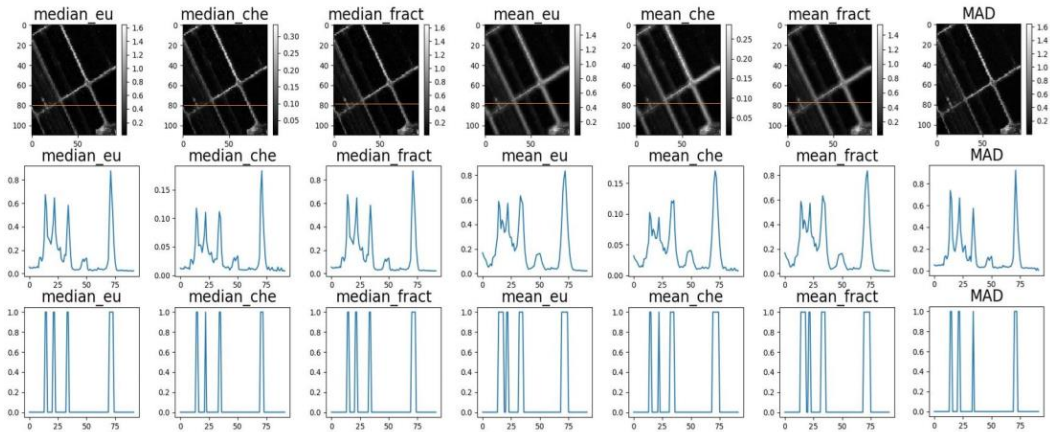

**Figure 4**. Edge maps generated using LSS (5×5 window) approach (top row). Horizontal profile of the edge strength map along the red line (middle row). Horizontal profile of the binarized edgemap (bottom row)

## 3. Experimental Study

### 3.1 Datasets

The proposed method is performed on airborne hyperspectral data sets. The first data was captured using the Airborne Visible and Infra-Red Imaging Spectrometer – Next Generation (AVIRIS-NG), of Jet Propulsion Laboratory (JPL), has been used for the ISRO-NASA airborne campaign onboard an ISRO-B200 aircraft. There are about 430 narrow continuous spectral bands in VNIR and SWIR regions in the range of 380-2510 nm at 5 nm interval with high SNR (>2000 @ 600 nm and >1000 @ 2200 nm) with pixel resolution 5m for flight altitude of 6km for a swath of 5km. Sixty noisy/water absorption bands were removed before the experiment. Fig.1(a) shows the AVIRIS-NG image captured over agricultural field at Muddur, Karnataka, India. Fig.1 (b) shows the spectra obtained from the image. The Salinas image was captured by the AVIRIS sensor over Salinas Valley, California, and with a spatial resolution of 3.7-meter per pixel. The image has 224 bands. Twenty water absorption bands were discarded before the experiment. Fig.1(e) shows the color composite of the Salinas image and the corresponding ground truth data. Airborne HSI data of Reno, NV, acquired on September 13, 2006, with the Prospectir sensor were also analysed in this study. Noisy bands and bands in those portions of the spectrum impacted by atmospheric water vapor were removed (322 bands).

### 3.2 Evaluation of Edge detection method

In general, three kinds of errors are involved in edge detection problem such as spurious responses(false positives), missing edges (false negatives) and displacements [16]. The FPs are usually related to texture and noises and FNs are linked with low contrast regions (due to image smoothing prior to edge extraction). Hence, proper evaluation measure to capture every kinds of errors in edge detection is desirable.

Finding the optimum ground truth solutions to evaluate the performance of an edge detection method is a challenging task. Some authors consider edge images generated from the standard edge detection methods to be the optimal solutions. However, different configuration of each method produce different results. In addition, error produced by the reference method may be considered as successful detection as well. Therefore, in this study, we have considered the actual edges (real situations) in an HSI on the basis of the context and the spatial feature [16] in the image. Fig. 1 shows the subset of the images that has been used for validating the performance of the edge detectors. Fig. 1 (c) indicates a tree (bright patch in the middle) and the agricultural land. Fig. 1 (d) shows the spatial subset of AVIRIS-NG image that consists of the cropping land (bright patch), the bare soil (dark patch) and the boundary between the patches. Fig. 1(e) shows the true colour composite of the Salinas image, the corresponding ground truth data and the white box shows the subset of the image used for testing the performance of edge detection methods. Different quality measures such as the number of missed edge pixels (miss count, MC), the number of falsely identified edge pixels (false alarm count, FAC) and Pratt's Figure of Merit (FOM) was employed to examine the performance of different edge detection methods [16]. Pratt's Figure of Merit is defined by,

$$FOM = \frac{1}{I_n}\sum_{i=1}^{I_A} \frac{1}{1+\alpha d^2} \quad (4)$$

where, $I_n = max(I_I, I_A)$. $I_I$ denotes the number of ideal edge points, $I_A$ indicates the number of actual edge points, $d$ is the displacement of the actual edge points from the the location of the ideal edge, and $\alpha$ is the scaling constant.

We have compared the proposed method with the other multi-spectral edge detection methods proposed in [4] and [6]. the horizontal profile of the edgemap (gray values) obtained from the proposed methods and the profile of the boolean edgemap generated from Otsu thresholding (performed on edgemap (gray values) derived using proposed methods). Sensitivity of the LSS over multivariate Gaussian noise were also examined by varying the noise intensity for different bands. In this case, different variance zero-mean Gaussian noise is added to each band, and the variance value being randomly selected from 0 to 0.05. [29].

### 3.3 Homogeneous area identification for Clustering and Classification

Extraction of more effective features in spatial domains to construct suitable classification models have already been applied to improve the classification accuracy [14][21][25].In HSI classification, spatial feature extraction is found more attractive through its effective exploitation of local textural information. Tarabalka et.al [26] suggested that the spatial information extracted from the neighborhood would be applied to improve the classification accuracy. They have also proposed a post-processing method (based on neighborhood similarity) to reduce the error in the final classification map. Moreover, Edge-based clustering techniques have also been applied to improve the segmentation by incorporating the discontinuities (or edges) in the image [10]. Building a thematic map based only on the pixel information often results in poor classification results [19]. In general, pixel based classifiers that can not cooperate the spatial variations at the edges in the image can't be used for generating accurate thematic maps. i.e. the pixel based classifier may produce errors at edges or spatial boundaries between classes [25].

Keeping such view, we have also examined the effect of removing the edges (or discontinuities) on the performance of clustering algorithms. At first, edges were extracted from the HSI using the LSS approach. Then the k-means clustering is applied on the image without considering the edge pixel vectors. Similarly, we applied LSS method as a pre-processing approach to reduce the error in the classification map generated using support vector machine (SVM). We considered SVM multi-class classifier with polynomial kernel (SVMP) to test the performance of the proposed classification framework. In our approach, classification consists of two steps: (1) mask the pixels which belong to edges using LSS method; (2) Apply the SVM classifier on the edge-masked image. This method can be used to mask the mixed pixels which are not part of the homogeneous patches. SVM kernel parameters were tuned iteratively for obtaining optimum results.

## 4 Results and Discussion

### 4.1 Edge detection in higher dimensional space

Qualitative and quantitative evaluation criteria were applied to investigate the effectiveness of the proposed LSS method. Performance of different distance measures and the corresponding statistical measures were analyzed using a subset of Salinas image. Fig. 2.1 shows the edge map obtained from the proposed LSS approach with 3×3 spatial window. It illustrates the results obtained from different spectral similarity measures and the corresponding statistical measures. Results show that the strength of the edge pixels generated using EU, CHE, EMD and FRACT are higher than the homogeneous areas. Similarly, edge strength map has been extracted using 7×7 spatial window. It shows that edge strength has been increased with increase in spatial window due to the inclusion of more neighborhood pixel vectors with edge information (also depends on spatial resolution). Moreover, the intensity difference between the edges (or heterogenous pixels) and the homogeneous areas were larger for median and MAD measures (computed from EU and FRACT). It shows that edgemap derived using EU, FRACT can be applied for distinguishing the edges and the homogeneous areas using binarization (e.g. Otsu thresholding) methods. Accordingly, Otsu thresholding [20] has been applied on the edge strength map to generate binarized edgemap. Fig. 2.3 illustrates the horizontal profile extracted from binarized image (generated from the edge strength map). This further shows that as window size increases, the distant pixels will have influence on the LSS. i.e. point statistical measures like mean may produce errors due to the presence of dissimilar spectra (or outliers) in the large spatial extent (window size = 5×5). Specifically, as window size increases, more distant pixels would be considered for finding the spectral similarity in the neighborhood. Therefore, the spatial extent may have more dissimilar pixel vectors (in region with large disparity) while computing spectral similarity between center pixel and the neighborhood. This effect has been examined with the image subset consists of a tree (size more than 5×5 spatial extent) at the middle of the cropping land (see Fig. 1 (c)). Fig. 2 shows that 7×7 window size identified the crown of the tree as a single spatial feature. However, 3×3 identified a homogeneous region at the middle of the tree patch because the small window consists of similar spectra within the neighborhood. In addition, median and MAD were able to identify the exact boundary of the feature as compared to the mean value of the local spectral similarity. This further proves that the statistical measure with less sensitivity over outliers can produce more accurate results.

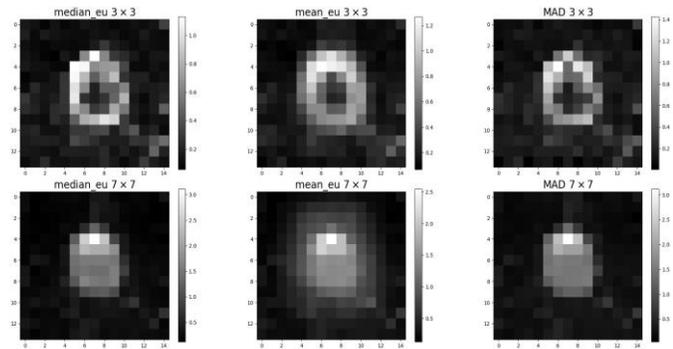

**Figure 5**: Edge strength map generated using different spatial window. 3×3 window (top row), 7×7 window (bottom row)

### 4.2 Edge detection in lower dimension

Fig. 6 shows the results of the DR methods applied on the test datasets. In this figure, homogeneous patches were identified by both the linear and non-linear DR methods, especially the LLE (see Fig. 6 (d)). However, these methods failed to enhance the exact boundary of the two different spatial regions. Therefore, representing the weak boundaries (or soft edges) in HSI using DR images may not produce accurate result. Moreover, post processing the HSI classification map using the DR images (as edge maps) methods may introduce errors in the final classification results. It is observed that extracting the edgemap using LSS after applying the DR may produce similar results (PCA and ISOMAP) as original data without DR (results not shown here). Moreover, edge strength map obtained from major components (components=3) of PCA shows more spurious edges as compared to PCA with more components (components= 115). Similar results were obtained from LLE method (see Fig. 4).

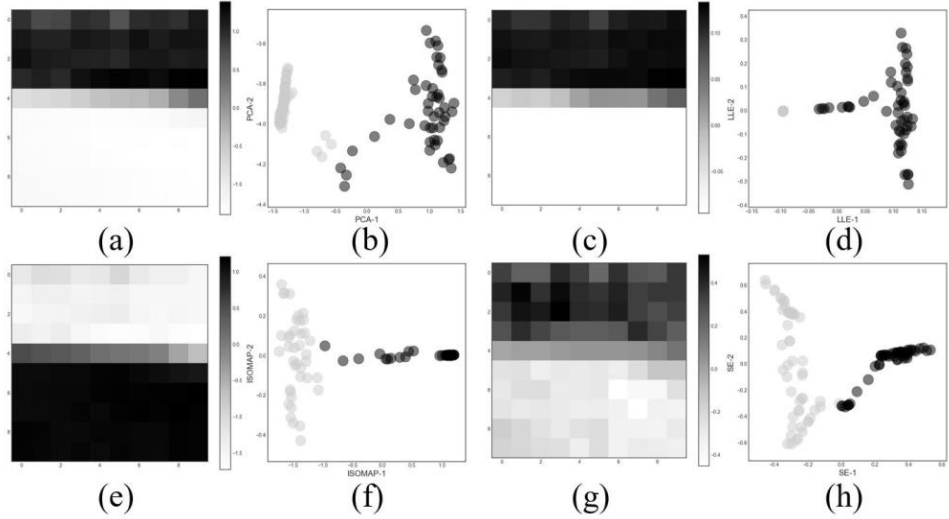

**Figure 6**: (a), (c), (e) and (f) show the grayscale image (see Fig.1 (d)) of the primary component of PCA, LLE, ISOMAP and SE respectively. (b), (d), (f) and (h) show the scatter plot of componenet-1 and component-2 of PCA, LLE, ISOMAP and SE respectively (black marks in the scatter plot represents black black pixels in the corresponding image)

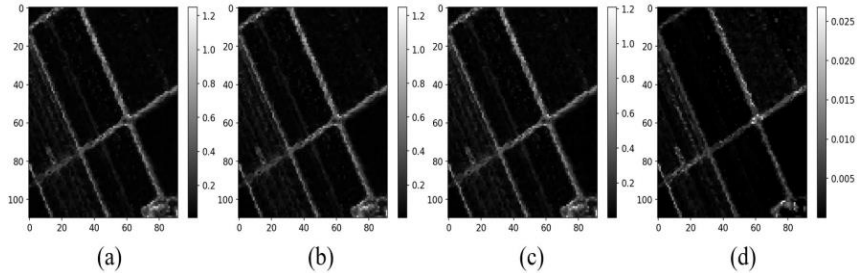

**Figure 7**: (a), (b), (c) and (d) show the edgemap derived from original image, PCA (15 components), PCA (3 components) LLE (3 components) respectively

## 4.3 Evaluation of LSS based edge detection

Different quality measures such as the number of missed edge pixels (miss count, MC), the number of falsely identified edge pixels (false alarm count, FAC) and Pratt's Figure of Merit (FOM) were applied to compare the performance of different methods. Table. 1 shows the performance of different edge detection methods on the test dataset shown in Fig.1 (d). This image subset consists of the narrow boundary (<2m) between bare soil and cropped land. However, the width of boundary (<2m) is smaller than the spatial resolution of the sensor (5m). This may lead to the mixing (or adjacency effect) of the adjacent features and finally produces week edges. Therefore, it is assumed that the method which is able to identify the week boundaries can be considered as an acceptable edge detection method for HSIs. Table. 1 also shows that the median-Eu, and median-FRACT had identified the edges correctly as compared to the other methods (as mentioned in [2]). Moreover, results show that performance of Median-Eu and Median-FRACT are better than [4] (FAC=3, MC=4) and [6] (FAC=10, MC=4). Figure 5 illustrates the performance of gradient (based on Euclidean distance) [4] edge detection methods on the test dataset shown in Fig. 1(d). Results (FOM, FAC and MC) demonstrate the poor performance of gradient based methods on the weak edges with mixed spectra.

Table 1: Comparison of different LSS approaches based on False Alarm Count, Miss Count and Pratt's Figure of Merit

|  | Median | | | Mean | | |
| --- | --- | --- | --- | --- | --- | --- |
|  | FAC | MC | FOM | FAC | MC | FOM |
| EU | 0 | 0 | 100 | 26 | 0 | 0 |
| COS | 1 | 0 | 0 | 24 | 0 | 0 |
| COR | 0 | 3 | 83 | 27 | 0 | 0 |
| CHE | 2 | 1 | 0 | 29 | 0 | 0 |
| EMD | 2 | 0 | 0 | 26 | 0 | 0 |
| FRACT | 0 | 0 | 100 | 26 | 0 | 0 |

Sensitivity of the proposed method against multivariate Gaussian noise has also been examined by adding different variance zero-mean Gaussian noise to each band, and the variance value being randomly selected from 0 to 0.05. Fig. 6 shows the edge strength map generated from the the noisy image by applying LSS with 3×3 and 7×7 spatial windows. Results show that 3×3 window based LSS method generated noisy patches in the edgemap. However, 7×7 window based approach produced clear edges especially using Median-Eu and median-FRACT. These observations describe the requirement of more neighborhood pixels (due to the spatial correlation of noise) to accurately compute the edges in noisy conditions. Moreover, Median based LSS illustrates its robustness to outliers as well as to multivariate Gaussian noise for estimating the local spectral variability.

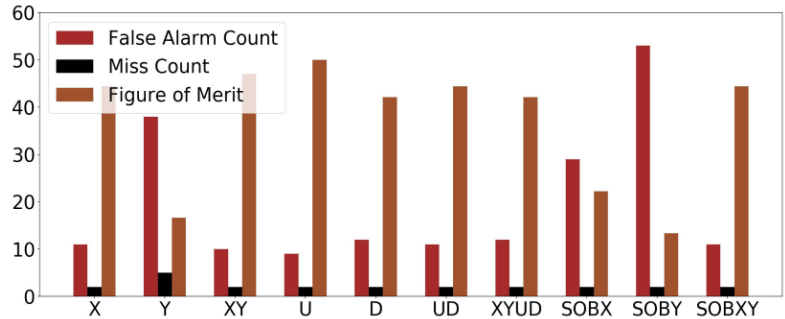

Figure 8: Comparison of proposed method with other Edge Detection Methods [4] ('gradient x', 'gradient y', 'gradient (x+y)/2','gradient up', 'gradient down', 'gradient', (up+down)/2','gradient (2x+2y+u+d)/6', 'sobel x', 'sobel y', 'sobelxy') based on FAC, MC and FOM

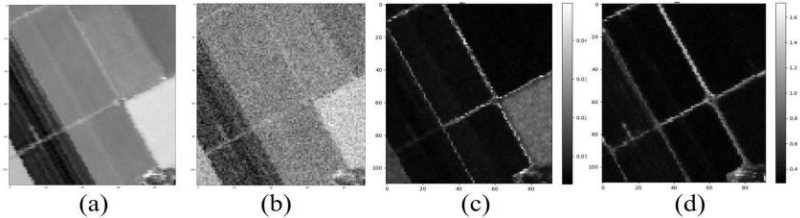

Figure 9: (a) Subset of the AVIRIS image (see Fig.1 (e)). (b) Image with multivariate Gaussian noise. (c) Edge strength map generated using LSS (3×3, Median-FRACT). (d) Edge strength map generated using LSS (7×7, Median-FRACT)

## 4.4 Edge detection applications for clustering and classification

Figure 7 illustrates the results obtained from SVM-polynomial and k-means clustering algorithms. Fig. 7 (c) shows that clustering of HSI without considering the edge information can produce the class map which is similar to the ground-truth image (see Fig. 7(b), Fig. 8 (b)). However, clustering of HSI without removing the edges produce erroneous results at the boundaries between different classes. This further shows that the pixel-vector based clustering that can not identify the spatial variations at the edges in the image due to the non-homogeneity at the edges or boundaries. Similarly, Fig. 7 (g) shows the classification result obtained from SVMP. In this figure, the boundaries of the agricultural land was classified as cropping land. However, LSS based approach was able to distinguish the spatial boundaries between similar classes and removed the errors in the final classification map due to the uncertainties contributed by the spatial boundaries. This may enhance the classifiers capabilities to discriminate small-area farms with narrow boundaries (see Fig. 7 (f)).

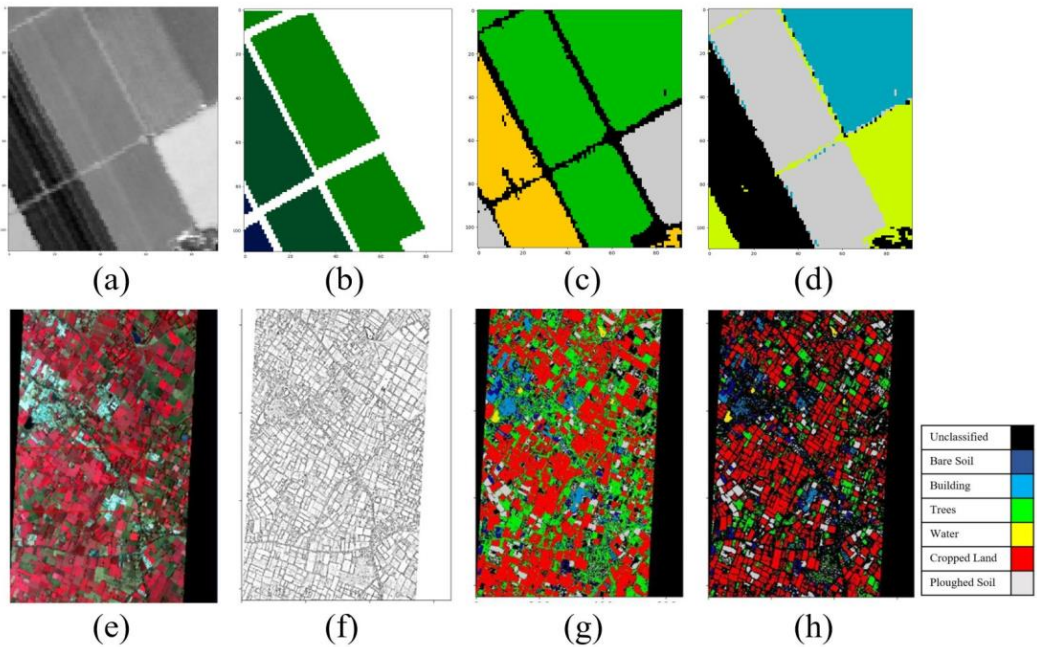

**Figure 10**: (a) Subset of the AVIRIS image (see Fig.1 (e)). (b) Groudtruth of the image. (c) K-means clustering applied on the image after removing the edges using LSS. (d) K-means clustering (3 classes) applied on the raw image. (e) Subset of the AVIRIS-NG image. (f) Edge strength map pf the AVIRIS-NG image. (g) SVMP classifier result. (h) SVMP classifier applied on the image after removing the edges using LSS

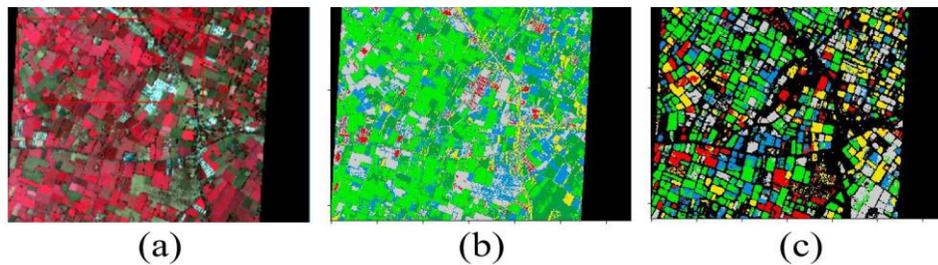

**Figure 11**: (a) Subset of the AVIRIS-NG image. (b) K-means clustering (6 classes) applied on the raw image. (c) K-means clustering applied on the image after removing the edges using LSS.

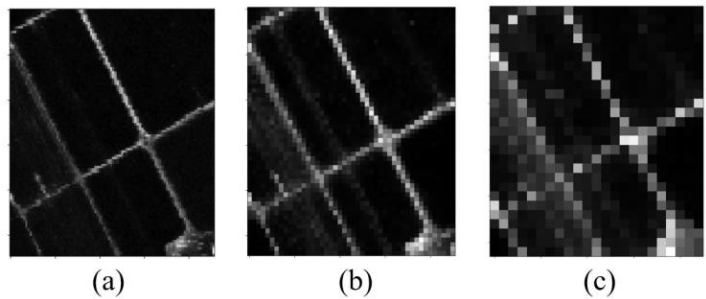

**Figure 12**: LSS applied on (a) raw image.(b) image downscaled by a factor of two.(c) image downscaled by a factor of four

Figure 10 illustrates the mutual-information (MI) between different bands and the spatial boundaries or edges. It shows that the variability in spectral response of edge features over different spectral bands (see Fig. 10). AVIRIS bands show comparatively better mutualinformation as compared to AVIRIS-NG image, this deviation in the performance may be coupled with the higher spatial resolution of AVIRIS (3.7m) or the wider boundaries between the agricultural plots or the availability of discriminant (spatially and spectrally) boundary features in the image. It further shows the importance of incorporating the spectral information distributed over the entire spectrum to identify the edges in HSI. We have also computed the MI for the spatially down-sampled (see Fig. 9) data. It was observed that as spatial resolution decreases, intensity of the edges also decreases This may lead to decrease in the mutual dependencies between the bands and the spatial boundaries. However, the proposed method was able to capture the edges from the down-sampled images as well. This further illustrates the potential of LSS to extract boundary features in low spatial resolution images as well.

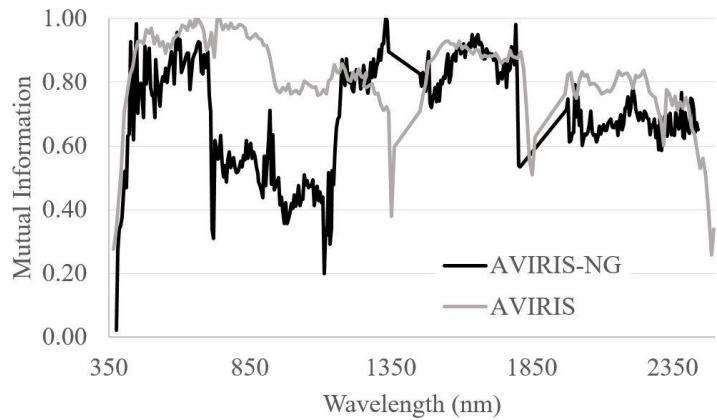

**Figure 13**: Normalized Mutual Information based band-sensitivity for detecting edges

Figure 11(c) and Fig. 11(d) show the results obtained from HySPADE (benchmark edge detection method for HSIs) [23] and the proposed method. Fig. 11(b) shows the 350th band of prospectir sensor with stripe noise. Therefore, effect of striping noise was appeared in HYSPADE result. However, LSS based method was able to remove the striping noise while extracting edge from the image. In addition, the analyst has to tune the edge detection method and modify the output planes in HySPADE individually using their histograms thus creating complexity for refining edge selection parameters of HySPADE [23]. Computational complexity of the HySPADE is also high (21 minutes for the complete Prospectir image) compared to the LSS (<2 minutes for the complete Prospectir image). Moreover, HySPADE may attempt to define an edge where one may not exist within a compositionally homogeneous patch in the imagery that is larger than the window size.

The key advantage of the proposed algorithm is that the simultaneous use of spatial and spectral information can reveal pixels with anomaly/unique spectrum (compared to other spectra with in a neighborhood) and enhances the homogeneity in the image. This is best demonstrated when the average value of the similarity matrix is applied to find the intensity of edges (mean statistics is prone to outliers). LSS may also be applied as a pre-processing approach before endmember extraction to mask the pixels with mixed responses or anomalies (the edges encompassing the mixed pixels in the scene have LSS responses greater than zero). i.e., mask out the edges prior to providing an HSI cube as an input to the endmember detection, clustering and classification algorithm can improve their results. However, as the number of pixel increases, the processing time of LSS also increases. Since the proposed method follows window based approach, great improvement in the computation time can be achieved by implementing the algorithm in parallel processing units.

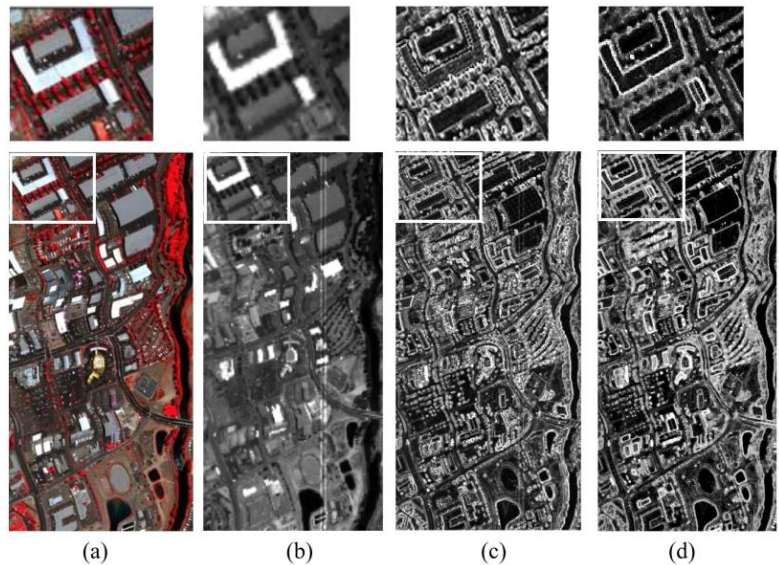

**Figure 14**: (a) False color composite of HSI image captured using prospectir sensor. (b) 350th band of prospectir sensor (with stripe noise) (c) Edgemap generated using HySPADE [23] (d) Edgemap generated using LSS (Median-FRACT). Top row shows the enlarged version of region within the white box

## 5. Conclusion

This study proposes an edge detection method for hyperspectral image classification based on vector similarity within a neighborhood. The proposed algorithm transforms the HSI image cube (within a spatial window) into a spectral similarity matrix by calculating the similarity between the center pixel-spectrum and the neighborhood spectra. As a result, high values in the similarity matrix represent the large changes (edges or disparity) in the local spatial extent. The final edge intensity is derived using order statistics or spatial convolution methods. The performance of the proposed algorithm has been benchmarked with state-of-the-art methods, including HySPADE [23], [4] and [6]. Experiments on a variety of datasets have validated the efficacy of the proposed algorithm for detecting edges in benchmark-HSIs under various levels of noise, spatial resolution, DR images etc. The experimental results also confirm that LSS can reduce the errors in clustering as well as classification results. In addition, the proposed algorithms outperform the traditional multichannel edge detectors in terms of both accuracy and the simplicity. However, the quality of detecting edges using LSS depends on the spectral-response (brightness) of the edge feature, window size of LSS, spectral similarity measure, convolution kernel or the order statistics. Therefore, further improvement may be achieved by adaptive selection of the window size and the optimum bands (with high edge information) which may contribute to multi-scale-edge detection in hyperspectral images.